\documentclass[10pt,journal,twocolumn]{IEEEtran}
\usepackage{graphicx}
\usepackage{amssymb}
\usepackage{amsmath}
\usepackage{subfigure}
\usepackage{multirow}
\usepackage{algorithm}
\usepackage{algpseudocode}
\usepackage{url}
\usepackage{booktabs}
  \usepackage{cite}
\ifCLASSINFOpdf
\else
\fi

\begin{document}
%
% paper title
% Titles are generally capitalized except for words such as a, an, and, as,
% at, but, by, for, in, nor, of, on, or, the, to and up, which are usually
% not capitalized unless they are the first or last word of the title.
% Linebreaks \\ can be used within to get better formatting as desired.
% Do not put math or special symbols in the title.
\title{Learning a Deep Part-based Representation by Preserving Data Distribution}

\author{Anyong~Qin,
        Zhaowei~Shang,
        Zhuolin~Tan,
       Taiping~Zhang,~\IEEEmembership{Member,~IEEE,}
        and~Yuan~Yan~Tang,~\IEEEmembership{Fellow,~IEEE}% <-this % stops a space
\thanks{Manuscript received Month, Year; revised Month, Year. The paper was supported by the National Natural Science Foundation of China (61906025, 61672114).}
\IEEEcompsocitemizethanks{
\IEEEcompsocthanksitem A. Qin and Z. Tan were with the School of Communication and Information Engineering,
Chongqing University of Posts and Telecommunications, Chongqing, China, 400065.\protect\\
E-mail: qinay@cqupt.edu.cn
\IEEEcompsocthanksitem Z. Shang and T. Zhang were with the College of Computer Science,
Chongqing University, Chongqing, China, 400030.\protect\\
\IEEEcompsocthanksitem Y.Y. Tang was with the Faculty of Science and Technology,
University of Macau, Macau, China, 999078.\protect\\
E-mail: yytang@umac.mo}% <-this % stops an unwanted space
}

% The paper headers
\markboth{Journal of \LaTeX\ Class Files,~Vol.~14, No.~8, August~2015}%
{Shell \MakeLowercase{\textit{et al.}}: Bare Demo of IEEEtran.cls for Computer Society Journals}

\IEEEtitleabstractindextext{%
\begin{abstract}
Unsupervised dimensionality reduction is one of the commonly used techniques in the field of high dimensional data recognition problems. The deep autoencoder network which constrains the weights to be non-negative, can learn a low dimensional part-based representation of data. On the other hand, the inherent structure of the each data cluster can be described by the distribution of the intraclass samples. Then one hopes to learn a new low dimensional representation which can preserve the intrinsic structure embedded in the original high dimensional data space perfectly. In this paper, by preserving the data distribution, a deep part-based representation can be learned, and the novel algorithm is called Distribution Preserving Network Embedding (DPNE). In DPNE, we first need to estimate the distribution of the original high dimensional data using the $k$-nearest neighbor kernel density estimation, and then we seek a part-based representation which respects the above distribution. The experimental results on the real-world data sets show that the proposed algorithm has good performance in terms of cluster accuracy and AMI. It turns out that the manifold structure in the raw data can be well preserved in the low dimensional feature space.
\end{abstract}

% Note that keywords are not normally used for peerreview papers.
}

\maketitle
\IEEEdisplaynontitleabstractindextext
\IEEEpeerreviewmaketitle

\section{Introduction}\label{sec:introduction}
\IEEEPARstart{U}{nsupervised} learning is a branch of machine learning algorithm that can learn inherent information from the unlabeled observation data. In many problems, the observation sample matrix is of very high dimension, for example, image and document recognition \cite{Turk1991Eigenfaces,Scott2010Indexing}. So it is infeasible to process the original data directly \cite{Duda2001Pattern}. One hopes then to learn a lower dimensional representation which still retains the inherent structure of the original data. In the last decade, a large number of linear and nonlinear methods for unsupervised dimensionality reduction have been proposed.

Singular Value Decomposition (SVD) is a common technique to analysis the multivariate data \cite{Klema1980The}, and has been successfully applied to face recognition \cite{Turk1991Eigenfaces} and document feature extracting \cite{Scott2010Indexing}. Principal Components Analysis (PCA) is a most popular unsupervised dimensionality reduction method. PCA embeds the original data into a low dimensional subspace that describes as much of the variance in the data as possible \cite{Pearson1901On,Jolliffe2002Principal}. Sammon mapping find a low dimensional representation, which can retain the pairwise distance of the high dimensional feature space\cite{Sammon1969A}. The Non-negative Matrix Factorization (NMF) can learn a low dimensional part-based representation of object because the two factors are constrained to be non-negative \cite{Lee1999Learning}. NMF has been shown the superiority in face recognition \cite{Li2001Learning} and document clustering \cite{Xu2003Document}. Isomap tries to retain the manifold (geodesic distance) between high dimensional data points, not to preserve the pairwise Euclidean distance \cite{Tenenbaum2000A}. Local Linear Embedding (LLE) assumes that the high dimensional data is a weight combination of his nearest neighbors, and then tries to reconstruct this local property in the low dimensional feature space \cite{Roweis2000Nonlinear}. Laplacian Eigenmaps attempts to retain the local manifold (Euclidean distance) in the low dimensional feature space. As a result, if the distance of two points in the high dimensional space is nearby, the corresponding low dimensional representations are also as close together as possible \cite{Belkin2001Laplacian}. Locality Preserving Projection (LPP) first constructs a neighborhood graph, and then find a optimal linear transformation which attempts to retain the local structure in a certain sense \cite{He2003Locality}. Similar to LPP, Neighborhood Preserving Embedding (NPE) also attempts to preserve the local neighborhood structure of the high dimensional data. However, the objective functions of LPP and NPE are totally different \cite{He2005Neighborhood}.

Multilayer autoencoders are the stacked feed-forward neural network and attempt to learn a complex nonlinear mapping function, which can automatically learn interesting feature from the input data \cite{Demers1992Non,Hinton2006Reducing}. The deep autoencoder can discover the inherent structure embedded in the high dimensional space by minimizing the mean squared error between the input data and the output of the network. For this reason, the deep autoencoder has been applied to various pattern recognition tasks, such as image, document \cite{Xie2016Unsupervised,Yang2017Towards} and hyperspectral land covers \cite{Luo2018Wavelet}.

Recently, the novel clustering methods, Deep Embedding Clustering (DEC) \cite{Xie2016Unsupervised} and Deep Clustering Network (DCN) \cite{Yang2017Towards} both make full use of the deep autoencoder to learn the latent representation, and achieve good performance. On the other hand, by constraining the non-negative property onto the two decomposition factors, NMF has been shown to be suitable for learning the parts of objects \cite{Lee1999Learning,Cai2011Graph}. For these reasons, the non-negative constraint autoencoder (NCAE) method is proposed to learn a part-based representation using deep autoencoder with non-negative constraint \cite{Hosseini2016}. Furthermore, \emph{Ayinde et al.} used the $l_1$ and $l_2$ regularization to induce the non-negative property. As a consequence, the learned feature is more interpretable and more suitable for clustering \cite{Ayinde2018Deep}.

However, these algorithms do not well capture or even ignore the nonlinear manifold structure imbedded in the high dimensional data space. In general, the data is sampling from the probability distribution that are near to a submanifold of the ambient space \cite{Cai2011Graph,Huang2011Subspaces}. Moreover, the nonlinear manifold structure of each class can be depicted by the distribution of the intraclass members. The intraclass samples are general located in a continuous high density area called density connected, while the different cluster are connected by some low density area called the border \cite{Ester1996,Tian2017Learning}. Preserving this manifold structure in the low dimensional feature space contributes to finding a clean and discriminative representation \cite{Qin2018Edge,Ding2018Spatial}.

In this paper, we demonstrate how to learn a meaningful data representation that explicitly considers the manifold structure. And we propose a novel dimensionality reduction technique, called Distribution Preserving Network Embedding (DPNE), which can learn a part-based representation using non-negative constraint autoencoder and simultaneously respect the distribution of original high dimensional data space. By estimating the distribution of high dimensional space, the manifold information embedded in the original high dimensional data space can be encoded. The goal of the proposed dimensionality reduction technique is to learn a meaningful data representation that can preserve the distribution of the original high dimensional data space as much as possible. As a consequence, two points that locate in a high density area are putting as close together as possible in the low dimensional feature space, and two points that are connected by low density area are far apart in the low dimensional representation.

\begin{figure}[t]
\centerline{\includegraphics[width=8cm,height=6cm]{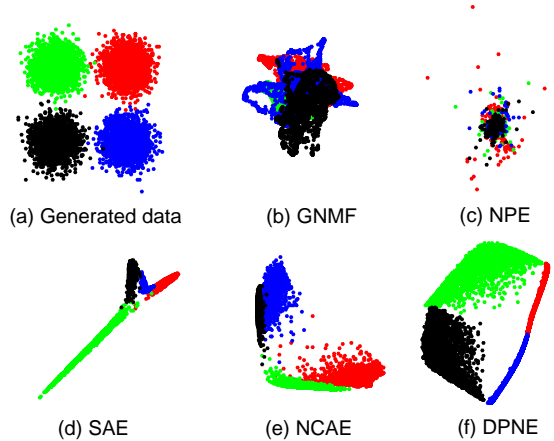}}
%\centerline{\includegraphics[width=0.5\textwidth,bb=0 0 420 315]{Vis_Gen.eps}}
\caption{Visualizing the synthetic data in a 2-dimensional space. The 2-dimensional friendly structure $h_i$ is a latent feature which can not be observed and we only have the high dimensional observation $x_i = \sigma(\omega^{(2)}\sigma(\omega^{(1)}h_i))$, where $\omega^{(1)}\in \mathbb{R}^{10\times 2}$ and $\omega^{(2)}\in \mathbb{R}^{100\times 10}$ are standard i.i.d. Gaussian distribution. (a) the generated k-means friendly 2-dimensional data, the recovered 2-dimensional data by (b) GNMF, (c) NPE, (d) SAE, (e) NCAE and (f) the proposed DPNE. For the three deep-based models (SAE, NCAE and DPNE), the list of layer size is $\{100, 50, 10, 2\}$.}
\label{Vis_Gen}
\end{figure}

A toy example that can be expected using the proposed DPNE is shown in Fig.~\ref{Vis_Gen}. The synthetic 2-dimensional data has friendly clustering structure in the 2-dimensional plane, and then the clustering structure is destroyed by mapping to a 100-dimensional space using a complex nonlinear function. We can see that the Graph regualrized Non-negative Matrix Factorization (GNMF) \cite{Cai2011Graph} and NPE fail to discover the clustering structure. Even though the visualizing results recovered by the deep-based SAE and NCAE model are superior to GNMF and NPE, the recovered 2-dimensional data are also not suitable for applying cluster method. As shown in Fig.~\ref{Vis_Gen}, the recovered 2-dimensional data obtained by the proposed method is most suitable for clustering analysis.

The main contributions of this paper include: 1) We use the $k$-nearest neighbor kernel density estimation to capture the inherent structure embedding in the high dimensional data space, and use the standard kernel density estimation to compute the distribution of the learned low dimensional feature space, and then minimize the inconsistency between the two distributions to achieve the goal of distribution preserving. 2) The proposed DPNE can learn a deep meaningful representation which preserves the distribution of high dimensional data space as much as possible. 3) The proposed unsupervised dimensionality reduction method tries to build a non-linear mapping function between the high dimensional data space and the low dimensional representation space by neural network. 4) We give the back-propagation rule of error for the distribution preserving network embedding model.

The remainder of the paper is organized as follows. Section \ref{Related_Work} describes the related work, i.e, sparse autoencoder (SAE) and non-negative constraint autoencoder (NCAE). In Section \ref{Proposed_Approach}, we describe the proposed distribution preserving network embedding (DPNE) along with the optimization scheme. The experimental results are shown and discussed in Section \ref{Experiments}. Finally, the conclusions are given in Section \ref{Conclusion}.

\section{Related Work}\label{Related_Work}
An autoencoder neural network is one unsupervised learning algorithm, which can automatically learn feature and then reconstruct its input at the output layer \cite{Vincent2008Extracting,Hinton1993Autoencoders}. It tries to learn two functions, i.e., encoder function $F(x)$ and decoder function $G(F(x))$. The encoder function $F(x)$ maps the input data to the feature space. Specifically, the computation of hidden representation is given by
\begin{eqnarray}
h^{(1)} = F(x)=\sigma(\omega^{(1)} x+ b^{(1)})
\label{encoding}
\end{eqnarray}
where $x$ is the input data, $\omega^{(1)}$ denotes the weight, $b^{(1)}$ represents the bias, and $\sigma (\cdot)$ is the activation function.
The decoder function $G(F(x))$ reconstructs the input data according to the hidden representation space. And the computation of reconstructing the input data is as follows,
\begin{eqnarray}
\hat{x} = G(h^{(1)})=\sigma(\omega^{(2)} h^{(1)}+ b^{(2)})
\label{decoding}
\end{eqnarray}
where $\omega^{(2)}$ denotes the weight and $b^{(2)}$ represents the bias. To optimize the parameters of the autoencoder neural network, i.e., $\Omega = \{\omega^{(1)},\omega^{(2)},b^{(1)},b^{(2)} \}$, the average reconstruction error is used as the objective function,
\begin{eqnarray}
\mathcal{O}(\Omega) = \min_{\Omega} \frac{1}{N}\sum_{i=1}^N \left\|x_i- \hat{x}_i\right\|_F^2
\label{obj_ae}
\end{eqnarray}
where $N$ represents the number of input data.

The stack autoencoder can learn semantically meaningful and well separated feature of input data \cite{Hinton2006Reducing,Vincent2010Stacked}. Thus, using the stack autoencoder contributes to discovering the latent structure embedded in the high dimensional space. Generally, a weight decay term is added to the Eq.~(\ref{obj_ae}) to help prevent overfitting \cite{Andrew2012Sparse}. And the final cost function of autoencoder is defined as
\begin{eqnarray}
\mathcal{O}(\Omega) =\frac{1}{N}\sum_{i=1}^N \left\|x_i- G(h_i^{(\frac{L}{2})})\right\|_F^2
+ \frac{\beta}{2} \sum_{l=1}^L\sum_{i=1}^{s_{l-1}}\sum_{j=1}^{s_l} (\omega_{ij}^{(l)})^2
\label{obj_ae1}
\end{eqnarray}
where $\Omega = \{\omega^{(l)},b^{(l)}\}$ denotes the parameters of the model with multiple hidden layers, the even $L$ denotes the number of layers, $\beta$ denotes the regularization parameter, $s_{l-1}$ and $s_l$ are the sizes of adjacent layers, and $h_i^{(\frac{L}{2})}$ denotes the output of $\frac{L}{2}$-th layer, i.e., the final low dimensional representation of sample $x_i$.

How to select the number of hidden units is crucial for the autoencoder method. When the number of hidden units is large, imposing some constraints on the hidden layers contributes to maintaining the proper number of active neuron \cite{Andrew2012Sparse}. Imposing a sparsity constraint on the hidden units, the autoencoder will still extract the latent structure hidden in the input data \cite{Hinton2006fast,Sch2007Efficient}. Enforcing the activation of hidden units to be near 0 is a common imposing the sparsity constraint method \cite{Lee2007Sparse,Vinod2009Object}. The average activation of the hidden unit $j$ is defined as
\begin{eqnarray}
\hat{p}_j = \frac{1}{N}\sum_{i=1}^{N}H_{i j}
\label{ave_act}
\end{eqnarray}
where $H_{i j}$ denotes the $j$-th element of the $i$-th hidden representation (i.e., $h_i$). By constraining the $\hat{p}_j = p$ ($p$ is a small positive value close to 0, for example, $p = 0.05$), the sparsity can be enforced \cite{Andrew2012Sparse}. Using the Kullback-Leibler divergence can achieve the constraint, i.e., $\hat{p}_j = p$
\begin{eqnarray}
KL(p\parallel \hat{p}) = \sum_{j=1}^{s_l} p~ log \frac{p}{\hat{p}_j} + (1-p)log \frac{1-p}{1-\hat{p}_j}
\label{KL_define}
\end{eqnarray}

To achieve sparsity constraint for the autoencoder neural network, an extra penalty term is added to the objective function Eq.~(\ref{obj_ae1}), and the final objective function of sparse autoencoder (SAE) can be written as
\begin{eqnarray}
\mathcal{O}(\Omega) =&& \frac{1}{N}\sum_{i=1}^N \left\|x_i- G(h_i^{(\frac{L}{2})})\right\|_F^2 + \alpha KL(p\parallel \hat{p}) \nonumber \\
&& + \frac{\beta}{2} \sum_{l=1}^L\sum_{i=1}^{s_{l-1}}\sum_{j=1}^{s_l} (\omega_{ij}^{(l)})^2
\label{obj_sae}
\end{eqnarray}

%NNSAE \cite{Lemme2012Online}

Many researches have demonstrated that utilizing the benefits of part-based representation with non-negative constraint can improve the performance of deep neural network \cite{Nguyen2013LearningPR,Lemme2012Online,Chorowski2015Learning,Hosseini2016}. To encourage the connecting weight $\omega^{(l)}$ to be non-negative, the regularization parameter in Eq.~(\ref{obj_sae}) is replaced as a quadratic function \cite{Nguyen2013LearningPR,Hosseini2016}. Thus, the objective function of non-negative constraint autoencoder (NCAE) can be written as
\begin{eqnarray}
\mathcal{O}(\Omega) =&& \frac{1}{N}\sum_{i=1}^N \left\|x_i- G(h_i^{(\frac{L}{2})})\right\|_F^2 + \alpha KL(p\parallel \hat{p}) \nonumber \\
&& + \frac{\beta}{2} \sum_{l=1}^L\sum_{i=1}^{s_{l-1}}\sum_{j=1}^{s_l} J(\omega_{ij}^{(l)})
\label{obj_ncae}
\end{eqnarray}
where
\begin{equation}\label{Non_constrained}
  J(\omega_{ij}^{(l)})=
  \begin{cases}
   (\omega_{ij}^{(l)})^2, &\mbox{$\omega_{ij}^{(l)} < 0$,}\\

   0, &\mbox{$\omega_{ij}^{(l)} > 0$.}

   \end{cases}
\end{equation}

\begin{figure}[t]
\centerline{\includegraphics[width=8cm,height=6cm]{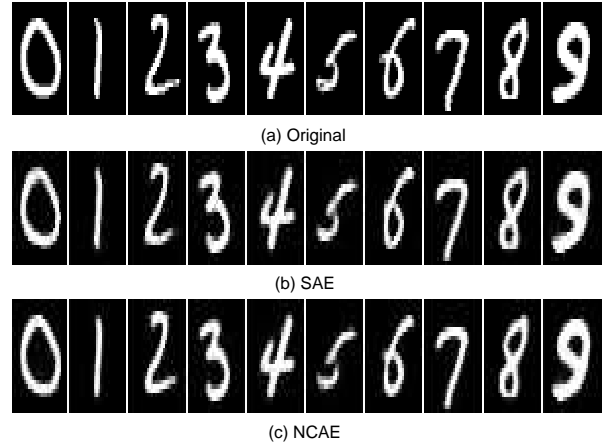}}
\caption{Reconstruction comparison of the MNIST digits with 500 receptive fields. (a) Original, (b) SAE (error = 3.53), (c) NCAE (error = 3.05).}
\label{reconstruction}
\end{figure}

Stacking the non-negative constraint autoencoder (NCAE) and a softmax classification layer would contribute to discovering the hidden structure of the high dimensional data, and improving the sparsity and reconstruction quality \cite{Hosseini2016}. Fig.~\ref{reconstruction} shows the randomly selected ten reconstruction MNIST digits with 500 receptive fields, as well as the average reconstruction errors. It can be seen that forcing the non-negative constraint onto the autoencoder to learn a part-based representation can result in more accurate reconstruction.
%$L_1L_2$AE \cite{Ayinde2018Deep}
%\begin{eqnarray}
%\mathcal{O}(\Omega) =&& \frac{1}{N}\sum_{i=1}^N \left\|x_i- G(h_i)\right\|_F^2 + \alpha KL(p\parallel \hat{p}) \nonumber \\
%&& + \sum_{l=1}^L\sum_{i=1}^{s_{l-1}}\sum_{j=1}^{s_l} J_{\ell_1/\ell_2}(\omega_{ij}^{(l)})
%\label{obj_l1l2ae}
%\end{eqnarray}
%where
%\begin{equation}
%  J_{\ell_1/\ell_2}(\omega_{ij}^{(l)})=
%  \begin{cases}
%   \beta_1 \Gamma(\omega_{ij}^{(l)})+ \frac{\beta_2}{2} \left\|\omega_{ij}^{(l)}\right\|^2, &\mbox{$\omega_{ij}^{(l)} < 0$,}\\
%
%   0, &\mbox{$\omega_{ij}^{(l)} > 0$.}
%
%   \end{cases}
%\end{equation}

% needed in second column of first page if using \IEEEpubid
%\IEEEpubidadjcol

\section{the Proposed Approach}\label{Proposed_Approach}
\subsection{The Standard Kernel Density Estimation}
Our early researches in manifold learning \cite{Tian2017Learning,Qin2018Edge} have demonstrated that preserving the nonlinear manifold structure of original data in the low dimensional space can be achieved by minimizing the inconsistence of two distributions \cite{J1965On}. Due to the lack of prior knowledge, it is desirable to use the popular nonparametric technique (such as kernel density estimation) to approximate the truthful distributions of data in high dimensional space and low dimensional space, respectively. We now briefly review the definition of the kernel density estimation. Given a finite sample set of $N$ points $x_i\in \mathbb{R}^M$, the estimator of the density $\hat{f}$ at the point $x$ is written as
\begin{equation}
\label{densityestimator}
f(x) = \frac{1}{N}\sum\limits_{i=1}^n \mathcal{K}_B(x-x_i)
\end{equation}
where $\mathcal{K}_B(\cdot)$ denotes the multivariate kernel function with a bandwidth matrix $B=(b_{ij})_{M\times N}$. For simplicity of presentation, we use one bandwidth parameter, i.e, $B=b^2I$, and
\begin{eqnarray}
\mathcal{K}_B(x)= | B |^{-1/2}\mathcal{K}(B^{-1/2}x) \nonumber
\end{eqnarray}
Then the Eq.~(\ref{densityestimator}) has the following well-known expression
\begin{equation}
\label{estimator}
f(x) = \frac{1}{N}\sum\limits_{i=1}^N \mathcal{K}_B(x-x_i) = \frac{1}{Nb^M}\sum\limits_{i=1}^N \mathcal{K}\left(\frac{x-x_i}{b}\right)
\end{equation}
In kernel density estimation, the kernel function is usually defined as the following special case
\begin{equation}
\mathcal{K}(x) = c\cdot \kappa(\|x\|^2)
\label{estimator_kernel}
\end{equation}
where $\kappa(\|x\|^2)$ is called $profile$ of the kernel and $c$ is a positive normalization constant\cite{Cheng1995Mean,Comaniciu2002Mean}. Substituting the Eq.~(\ref{estimator_kernel}) into Eq.~(\ref{estimator}), the estimator of density $\hat{f}$ has the form
\begin{equation}
\label{finalform}
f(x) = \frac{c}{Nb^M}\sum\limits_{i=1}^N \kappa\left(\left\|\frac{x-x_i}{b}\right\|^2\right)
\end{equation}
On the other hand, the marginal density function is typically calculated by summing the joint density function, so the density estimation can also be written as
\begin{eqnarray}
f(x) = \sum_{i=1}^N f(x, x_i)
\end{eqnarray}
Furthermore, the density estimation is decomposed in a probabilistic way as
\begin{eqnarray}
f(x) = \sum_{i=1}^N f(x|x_i)P(x_i),
\end{eqnarray}
where $P(x_i)$ is a priori probability. If we assume the given data independent and identically distributed (i.i.d), i.e, $P(x_i)=1/N$, it is easy to show that the kernel function is proportional to the conditional density, $\mathcal{K}_B(x-x_i)\propto f(x|x_i)$. According to the density estimation theory, the densities of the given data and the low dimensional feature can be computed with the following formula, respectively.
\setlength{\arraycolsep}{0.0em}
\begin{eqnarray}
&&\label{high}f(x) = \sum\limits_{i=1}^N f(x|x_i) = \frac{c}{Nb_x^M}\sum\limits_{i=1}^N \kappa\left(\left\|\frac{x-x_i}{b_x}\right\|^2\right)\\
&&\label{low}g(h) = \sum\limits_{i=1}^N g(h|h_i)= \frac{c}{Nb_h^D}\sum\limits_{i=1}^N \kappa\left(\left\|\frac{h-h_i}{b_h}\right\|^2\right)
\end{eqnarray}
where $D$ is the dimension of the learned representation $h_i$. The parameters $b_x$ and $b_h$ should satisfy the nonlinear equation $\sum\limits_{i=1}^N \mathcal{K}_B(x-x_i) log_2(\mathcal{K}_B(x-x_i))= log_2(t)$ and $t$ is set to 20 \cite{Tian2017Learning}. To preserve the distribution of original data in the low dimensional feature space, many criteria can be used to measure the inconsistency, for example the Kullback-Leibler divergence criterion.
\begin{eqnarray}
\label{obj_dpe}
f(x)\log{\frac{f(x)}{g(h)}}
\end{eqnarray}

\subsection{$k$-Nearest Neighbour Kernel Density Estimation}
We use the kernel density estimation to capture the manifold structure. Thus, the key problem for the proposed method is how to estimate the density of samples. However, the standard kernel density estimation has poor performance in the high dimensional space. The $k$-nearest neighbor kernel density estimation is a special case of the standard kernel density estimation with the local variables. According to the number of samples in a local region, we can smooth this estimator to obtain the more approximate density \cite{Rosenblatt1979,Orava2011K}. The $k$-nearest neighbor kernel density estimation described in \cite{Rosenblatt1979,Orava2011K} can be written as
\begin{equation}
\label{Knn_estimator}
f(x,k) = \frac{c}{Nd_x^M}\sum\limits_{i=1}^N \mathcal{K}\left(\frac{x-x_i}{d_x}\right)
\end{equation}
where $d_x=d(x)$ is a Euclidean distance between $x$ and the $k$-th nearest neighbor of $x$ among $x_j$'s,
\begin{equation}
\label{Knn}
d(x) = min(k,\{ \lvert x-x_j \rvert , j=1,2,\cdots,N \}),
\end{equation}
where $min(k,A)$ is the $k$-th smallest element of the set $A$. So the density of the given data is
\begin{eqnarray}
\label{high1}f(x) = \sum\limits_{i=1}^N f(x|x_i) = \frac{c}{Nd_x^M}\sum\limits_{i=1}^N \kappa\left(\left\|\frac{x-x_i}{d_x}\right\|^2\right)
\end{eqnarray}

\subsection{DPNE}
If we assume the learning low dimensional feature preserves the distribution of the given data very well, it follows that the conditional density of the given data at point $x_i$ ($x_i \in X$) and the conditional density of the corresponding low dimensional representation at point $h_i$ ($h_i\in H$) will be equal. In other words, the goal of distribution preserving is to find a lower dimensional representation $H$ that minimizes the inconsistency between $f(x_i|x_j)$ and $g(h_i|h_j)$ for all ($x_i,h_i$). To achieve the goal, we try to minimize the inconsistency between any $f(x_i|x_j)$ and $g(h_i|h_j)$,
\begin{eqnarray}
\label{obj1}
\sum_{i=1}^N\sum_{j=1}^Nf(x_i|x_j)\log{\frac{f(x_i|x_j)}{g(h_i|h_j)}}
\end{eqnarray}

This paper does not focus on the choice of kernel function $\kappa(\cdot)$ in the step of kernel density estimation. So we employ the Gaussian kernel and Cauchy kernel to estimate the distributions of high dimensional space (Eq.~(\ref{high1})) and low dimensional space (Eq.~(\ref{low})), respectively. The Cauchy kernel is a long-tailed kernel and has the ability to alleviate the crowding problem in the low dimensional space \cite{Tian2017Learning,Qin2018Edge}. By minimizing Eq.~(\ref{obj1}), we expect that if two samples $x_i$ and $x_j$ are close, the low dimensional representations $h_i$ and $h_j$ are also close to each other, and vice versa. Thus, as an extra penalty term, we add Eq.~(\ref{obj1}) to the Eq.~(\ref{obj_ncae}). The greedy layer-wise trained NCAE model is used to initialize the proposed DPNE network, and the resulting network is not imposed sparsity constraint in the fine-tuning stage. The final cost function of the proposed DPNE network is as follows
\begin{eqnarray}
\mathcal{O}(\Omega,h) =&& \frac{1}{N}\sum_{i=1}^N \left\|x_i- G(h_i^{(\frac{L}{2})})\right\|_F^2  \nonumber \\
&& + \frac{\beta}{2} \sum_{l=1}^L\sum_{i=1}^{s_{l-1}}\sum_{j=1}^{s_l} J(\omega_{ij}^{(l)}) \nonumber\\
&& + \gamma \sum_{i=1}^N\sum_{j=1}^Nf(x_i|x_j)\log{\frac{f(x_i|x_j)}{g(h_i^{(\frac{L}{2})}|h_j^{(\frac{L}{2})})}} \nonumber\\
 = && \mathcal{O}_{rec} + \frac{\beta}{2}\mathcal{O}_{reg} + \gamma\mathcal{O}_{dp}
\label{obj_dpne}
\end{eqnarray}
where
\begin{equation}\label{Non_constrained}
  J(\omega_{ij})=
  \begin{cases}
   \omega_{ij}^2, &\mbox{$\omega_{ij} < 0$,}\\

   0, &\mbox{$\omega_{ij} > 0$.}

   \end{cases}
\end{equation}
and $\gamma$ controls the penalty term facilitating distribution preserving.

\subsection{Optimization}
Our goal is to minimize $\mathcal{O}(\Omega,h)$ (i.e., Eq.~(\ref{obj_dpne})) as a function of $\Omega$, to obtain the part-based representation which can well preserve the distribution of the given data. The gradient descent method can be used to optimize the objective Eq.~(\ref{obj_dpne}). We first calculate the partial derivative of $\mathcal{O}_{reg}$, (i.e., $\frac{\partial \mathcal{O}_{reg}}{\partial \omega^{(l)}}$) as follows
%\begin{eqnarray}
%\label{gradient_reg1}
%\frac{\partial \mathcal{O}_{reg}}{\partial \omega_{ij}^{(l)}} = \sum_{l=1}^L\sum_{i=1}^{s_l}\sum_{j=1}^{s_{l+1}} J'(\omega_{ij}^{(l)})
%\end{eqnarray}

\begin{equation}\label{gradient_reg2}
  \frac{\partial \mathcal{O}_{reg}}{\partial \omega_{ij}^{(l)}} =
  \begin{cases}
   2\omega_{ij}^{(l)}, &\mbox{$\omega_{ij}^{(l)} < 0$,}\\

   0, &\mbox{$\omega_{ij}^{(l)} > 0$.}

   \end{cases}
\end{equation}
where $l=1,2,\cdots,L$.

Then we consider the $\frac{\partial \mathcal{O}_{rec}}{\partial \omega^{(L)}}$ and it can be rephrased as follows
\begin{eqnarray}
\label{gradient_rec1}
\frac{\partial \mathcal{O}_{rec}}{\partial \omega^{(L)}} = \frac{\partial \mathcal{O}_{rec}}{\partial z_i^{(L)}} \cdot \frac{\partial z_i^{(L)}}{\partial \omega^{(L)}}
\end{eqnarray}

According to $ z_i^{(l)} = \omega^{(l)} h_i^{(l-1)}+ b^{(l)}$, we have the form of partial derivative as follows

\begin{eqnarray}
\label{gradient_acti}
\frac{\partial z_i^{(l)}}{\partial \omega^{(l)}} = h_i^{(l-1)}
\end{eqnarray}
where $l = 1,2,\cdots, L$, and $h^{(0)} = x$ denotes the input data.

The first term in Eq.~(\ref{gradient_rec1}) is easy to calculate
\begin{eqnarray}
\label{gradient_bp1}
\frac{\partial \mathcal{O}_{rec}}{\partial z_i^{(L)}} = (\hat{x}_i - x_i)\star \sigma '(z_i^{(L)})
\end{eqnarray}
where $\star$ denotes the element-wise product operator. Using the back-propagation rule, we can iteratively obtain the $\frac{\partial \mathcal{O}_{rec}}{\partial z_i^{(l)}}$, i.e.,

\begin{eqnarray}
\label{gradient_bp2}
\frac{\partial \mathcal{O}_{rec}}{\partial z_i^{(l)}} = (\omega^{(l+1)} \frac{\partial \mathcal{O}_{rec}}{\partial z_i^{(l+1)}})\star \sigma '(z_i^{(l)})
\end{eqnarray}
where $l = L-1, L-2, \cdots, 1$.

Thus, the $\frac{\partial \mathcal{O}_{rec}}{\partial \omega^{(l)}}$ can be iteratively obtained, where $l = 1,2,\cdots, L$. To update the weight of decoder part (i.e., $l=\frac{L}{2}+1,\frac{L}{2}+2,\cdots,L$), the gradient of Eq.~(\ref{obj_dpne}) has the following form
\begin{eqnarray}
\label{gradient1}
\frac{\partial \mathcal{O}}{\partial \omega^{(l)}} = \frac{\partial \mathcal{O}_{rec}}{\partial \omega^{(l)}} + \frac{\beta}{2}\frac{\partial \mathcal{O}_{reg}}{\partial \omega^{(l)}}
\end{eqnarray}

To update the weight of encoder part (i.e., $l=1,2,\cdots,\frac{L}{2}$), the gradient of Eq.~(\ref{obj_dpne}) has the similar form
\begin{eqnarray}
\label{gradient2}
\frac{\partial \mathcal{O}}{\partial \omega^{(l)}} = \frac{\partial \mathcal{O}_{rec}}{\partial \omega^{(l)}} + \frac{\beta}{2}\frac{\partial \mathcal{O}_{reg}}{\partial \omega^{(l)}}+\gamma \frac{\partial \mathcal{O}_{dp}}{\partial \omega^{(l)}}
\end{eqnarray}

%\begin{eqnarray}
%\label{gradient_rec2}
%\frac{\partial \mathcal{O}_{rec}}{\partial \omega_{ij}^{(l)}} = \frac{\partial \mathcal{O}_{rec}}{\partial \hat{X}} \cdot \frac{\partial \hat{X}}{\partial \omega_{ij}^{(L)}}
%\end{eqnarray}
We now only need to calculate the partial derivative $\frac{\partial \mathcal{O}_{dp}}{\partial \omega^{(l)}}$. And it can be rephrased as follows
\begin{eqnarray}
\label{gradient_dp}
\frac{\partial \mathcal{O}_{dp}}{\partial \omega^{(l)}} = \frac{\partial \mathcal{O}_{dp}}{\partial h_i^{(\frac{L}{2})}} \cdot
 \frac{\partial h_i^{(\frac{L}{2})}}{\partial \omega^{(l)}}
\end{eqnarray}
where the first term $\frac{\partial \mathcal{O}_{dp}}{\partial h_i^{(\frac{L}{2})}}$ is given by
\begin{eqnarray}
\label{gradient_dp1}
\frac{\partial \mathcal{O}_{dp}}{\partial h_{ i}^{(\frac{L}{2})}} = && \frac{c}{Nb_{h^{(\frac{L}{2})}}^{K}} \sum\limits_{j=1}^N \left(f(x_i|x_j) - g(h_i^{(\frac{L}{2})}|h_j^{(\frac{L}{2})})\right) \nonumber\\
&&\cdot (h_i^{(\frac{L}{2})}-h_j^{(\frac{L}{2})})\kappa '\left(\left\|\frac{h_i^{(\frac{L}{2})}-h_j^{(\frac{L}{2})}}{b_h^{(\frac{L}{2})}}\right\|^2\right)
\end{eqnarray}

The second term $\frac{\partial h_i^{(\frac{L}{2})}}{\partial \omega^{(l)}}$ can be rephrased as follows
\begin{eqnarray}
\label{gradient_dp2}
\frac{\partial h_i^{(\frac{L}{2})}}{\partial \omega^{(l)}}&& =  \frac{\partial h_i^{(\frac{L}{2})}}{\partial z_i^{(l)}} \cdot \frac{\partial z_i^{(l)}}{\partial \omega^{(l)}} \nonumber \\
&& = \frac{\partial h_i^{(\frac{L}{2})}}{\partial z_i^{(l)}} \cdot h_i^{(l-1)}
\end{eqnarray}

Note that the $\frac{L}{2}$-th layer of the proposed DPNE has a linear activation function to make the final data representation maintain full information \cite{Vincent2010Stacked,Xie2016Unsupervised}. Specifically, the computation of the $\frac{L}{2}$-th layer is given by
\begin{eqnarray}
\label{linear_act}
h_i^{(\frac{L}{2})} = z_i^{(\frac{L}{2})} = \omega^{(\frac{L}{2})} h_i^{(\frac{L}{2}-1)}+ b^{(\frac{L}{2})}
\end{eqnarray}

Then we have
\begin{eqnarray}
\label{gradient_dp_bp1}
\frac{\partial h_i^{(\frac{L}{2})}}{\partial z_i^{(\frac{L}{2})}} = \textbf{1}
\end{eqnarray}

Based on back-propagation rule, we have the following mathematical form of the partial derivative
\begin{eqnarray}
\label{gradient_dp_bp2}
\frac{\partial h_i^{(\frac{L}{2})}}{\partial z_i^{(l)}} = (\omega^{(l+1)} \frac{\partial h_i^{(\frac{L}{2})}}{\partial z_i^{(l+1)}}) \star  \sigma '(z_i^{(l)})
\end{eqnarray}
where $l = \frac{L}{2}-1, \frac{L}{2}-2, \cdots, 1$.

By the same token, the gradient of $\mathcal{O}(\Omega,h)$ with respect to the bias of reconstruction layer $b^{(l)}$ (i.e., $l = L, L-1, \cdots, \frac{L}{2}+1$) is written as

\begin{eqnarray}
\label{gradient_bias1}
\frac{\partial \mathcal{O}}{\partial b^{(l)}} = \frac{\partial \mathcal{O}_{rec}}{\partial z_i^{(l)}}
\end{eqnarray}
and the bias of mapping layer $b^{(l)}$ (i.e., $l = \frac{L}{2},\frac{L}{2}-1,\cdots,1$) as
\begin{eqnarray}
\label{gradient_bias2}
\frac{\partial \mathcal{O}}{\partial b^{(l)}} = \frac{\partial \mathcal{O}_{rec}}{\partial z_i^{(l)}} + \frac{\partial \mathcal{O}_{dp}}{\partial h_{ i}^{(\frac{L}{2})}} \frac{\partial h_i^{(\frac{L}{2})}}{\partial z_i^{(l)}}
\end{eqnarray}

\begin{figure}[t]
\centerline{\includegraphics[width=9cm,height=7cm]{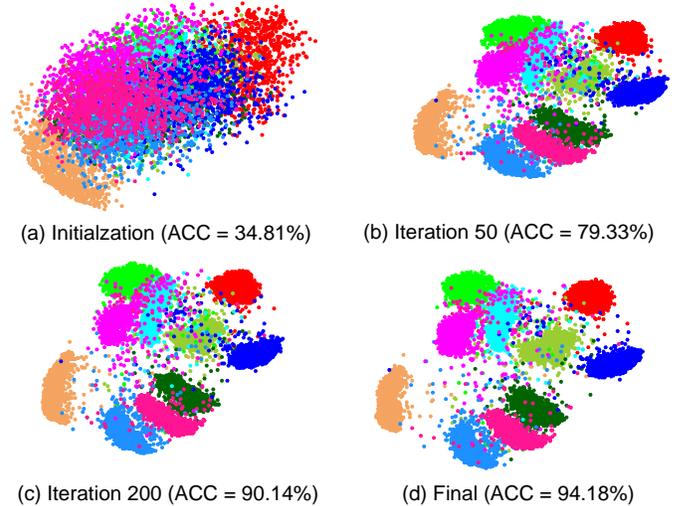}}
\caption{The DPNE on the MNIST data set. (a) The 2-dimensional embedding initialized by the NCAE. (b) The embedding $H$ after 50 iterations. (c) The embedding $H$ after 200 iterations. (d) The final 2-dimensional embedding produced by the proposed DPNE. The corresponding cluster accuracy (ACC) based on the 2-dimensional embedding is shown in the bracket. The ground truth classes are coded by color.}
\label{optimization_process}
\end{figure}

The weights and basis $\Omega$ are greedy layer-wise initialized using the non-negative constraint autoencoder (NCAE) model \cite{Hosseini2016}. Each pair of corresponding encoding and decoding layers is initialized sequentially. Then our DPNE is fine-tuned end-to-end according to the proposed objective function Eq.~(\ref{obj_dpne}). If the stop criterion meets, the DPNE terminates and outputs the low dimensional embedding. The proposed DPNE is summarized in Algorithm \ref{DPNE}. As shown in Fig.~\ref{optimization_process}, the embedding is gradually optimized to preserve the inherent structure of the data. It turns out that our DPNE not only capture the manifold structure embedded in the high dimensional space, but also well preserve the structure in the low dimensional feature space.
%The framework of the proposed DPNE ?, refer to \cite{Wang2016Structural} Figure 2.

\renewcommand{\algorithmicrequire}{\textbf{Input:}} % Use Input in the format of Algorithm
\renewcommand{\algorithmicensure}{\textbf{Output:}} % Use Output in the format of Algorithm
\begin{algorithm}[h]
\caption{Distribution Preserving Network Embedding (DPNE)}
\label{DPNE}
\begin{algorithmic}[1]
\Require
Original data $x_i \in X$.
\Ensure
low dimensional embedding $h_i^{(\frac{L}{2})}\in H$.
\State Initialize $\beta = 0.003$, $\gamma = 100$, learning rate $\eta = 0.1$, $L = 8$, maximum of number of iterations $maxiter = 400$, the dimension of the part-based representation $D = 10$ and the number of nearest neighbors $k = 10$.
\State Compute the original data distribution $f(x_i)$ (using Eq.~(\ref{high1})).
\For {all layers}
\State Use two layer NCAE to initialize the corresponding weights and bias.
\EndFor
\State Obtain the initial embedding $h_i^{(\frac{L}{2})}$.
\For {$iter = 1,2,\cdots, maxiter$}
\State Compute the distribution of low dimensional embedding $g(h_i^{(\frac{L}{2})})$ (using Eq.~(\ref{low})).
\State Initialize $\triangle \omega^{(l)} = 0$, $\triangle b^{(l)} = 0$
\For  {$i = 1,2, \cdots, N$}
\State Use the back-propagation rule to compute $\frac{\partial \mathcal{O}_{rec}}{\partial \omega^{(l)}}$, $\frac{\partial \mathcal{O}_{dp}}{\partial \omega^{(l)}}$, and $\frac{\partial \mathcal{O}}{\partial b^{(l)}}$
\If {$l = L, L-1, \cdots, \frac{L}{2}+1$}
\State $\triangle \omega^{(l)} = \triangle \omega^{(l)} + \frac{\partial \mathcal{O}_{rec}}{\partial \omega^{(l)}}$
\State $\triangle b^{(l)} = \triangle b^{(l)} + \frac{\partial \mathcal{O}}{\partial b^{(l)}}$
\ElsIf {$l = \frac{L}{2},\frac{L}{2}-1,\cdots,1$}
\State $\triangle \omega^{(l)} = \triangle \omega^{(l)} + \frac{\partial \mathcal{O}_{rec}}{\partial \omega^{(l)}} + \gamma \frac{\partial \mathcal{O}_{dp}}{\partial \omega^{(l)}}$
\State $\triangle b^{(l)} = \triangle b^{(l)} + \frac{\partial \mathcal{O}}{\partial b^{(l)}}$
\EndIf
\EndFor
\State // Use the gradient descent method to update the parameters $\omega^{(l)}$ and $b^{(l)}$.
\State $\omega^{(l)} = \omega^{(l)} - \eta [(\frac{1}{N}\triangle \omega^{(l)}) + \frac{\beta}{2} \frac{\partial \mathcal{O}_{reg}}{\partial \omega^{(l)}}]$
\State $b^{(l)} = b^{(l)} - \eta [\frac{1}{N}\triangle b^{(l)}]$
\State Obtain the final network embedding $h_i^{(\frac{L}{2})} = F(x_i)$
\EndFor
\end{algorithmic}
\end{algorithm}

\section{Experiments}\label{Experiments}
In this paper, we compare the representation space obtained by the proposed DPNE to the classic and deep model algorithms, i.e, k-means++ \cite{Arthur2007k}, Local Discriminant Models and Global Integration (LDMGI) \cite{Yang2010Image}, Graph Degree Linkage (GDL) \cite{Zhang2012Graph}, Spectral Embedded Clustering (SEC) \cite{Nie2011Spectral}, Deep Embedding Clustering (DEC) \cite{Xie2016Unsupervised}, Deep Clustering Network (DCN) \cite{Yang2017Towards}, Robust Continuous Clustering (RCC) \cite{Shah2017Robust}, Sparse Autoencoder (SAE) \cite{Andrew2012Sparse}, Non-negative Constraint Autoencoder (NCAE) \cite{Hosseini2016} and GNMF \cite{Cai2011Graph}. For the SAE, NCAE and DPNE model, we use the k-means method to cluster the low dimensional representation in our paper. The cluster accuracy (ACC) and the adjusted mutual information (AMI) are employed to evaluate the performance of these different algorithms. The default parameters for the compared algorithms are used. The input parameters of our algorithm are as follows: regularization parameters $\beta = 0.003$ and $\gamma = 100$, learning rate $\eta = 0.1$, the number of layers $L = 8$, the number of iterations $maxiter = 400$ and the number of nearest neighbors $k = 10$, as shown in Table \ref{Parameter}. Same as the DEC and DCN, the list of the layer size is $M-500-500-2000-D$ ($M\gg D$), where $M$ is the dimension of the original high dimensional data space and $D$ is the dimension of the learned feature space \cite{Xie2016Unsupervised,Yang2017Towards}. For all methods, we repeat ten times to obtain reliable and stable results of each data set.
\begin{table}[!t]
%\scriptsize
%\renewcommand\arraystretch{1.5}
\caption{Parameter settings of each algorithm.}
\label{Parameter}
\centering
        \begin{tabular}{ccccc}
        \hline
         Parameters   &  SAE & NCAE &  GNMF &  DPNE \\
         \hline
         Sparsity penalty ($\alpha$) & 3  & 3 & -  & - \\
         Sparsity parameter (p) & 0.05  & 0.05  & -  & - \\
         Weight decay penalty ($\beta$) & 0.003  & -  & -  & - \\
         Nonnegativity constraint ($\beta$) & -  & 0.003  & -  & 0.003 \\
         Regularization parameter ($\gamma$) & -  & -  & 100  & 100 \\
         Maximum No. of Iterations & 400  & 400  & 400  & 400 \\
        \hline
        \end{tabular}
\end{table}

\begin{figure*}[t]
\centerline{\includegraphics[width=16cm,height=6cm]{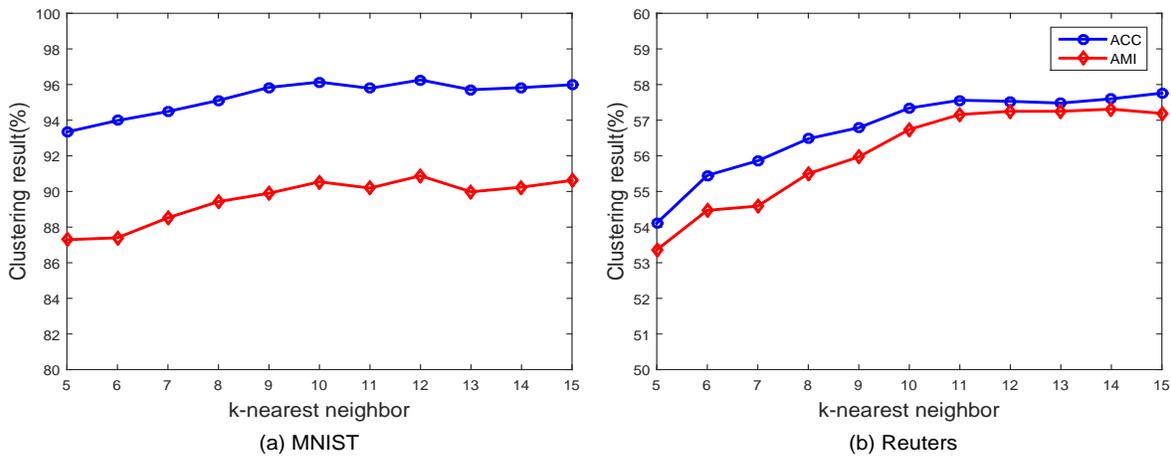}}
\caption{The clustering performance of DPNE vs. the number of nearest neighbors $k$.}
\label{nearest_neighbor}
\end{figure*}

\begin{figure*}[t]
\centerline{\includegraphics[width=18cm,height=7cm]{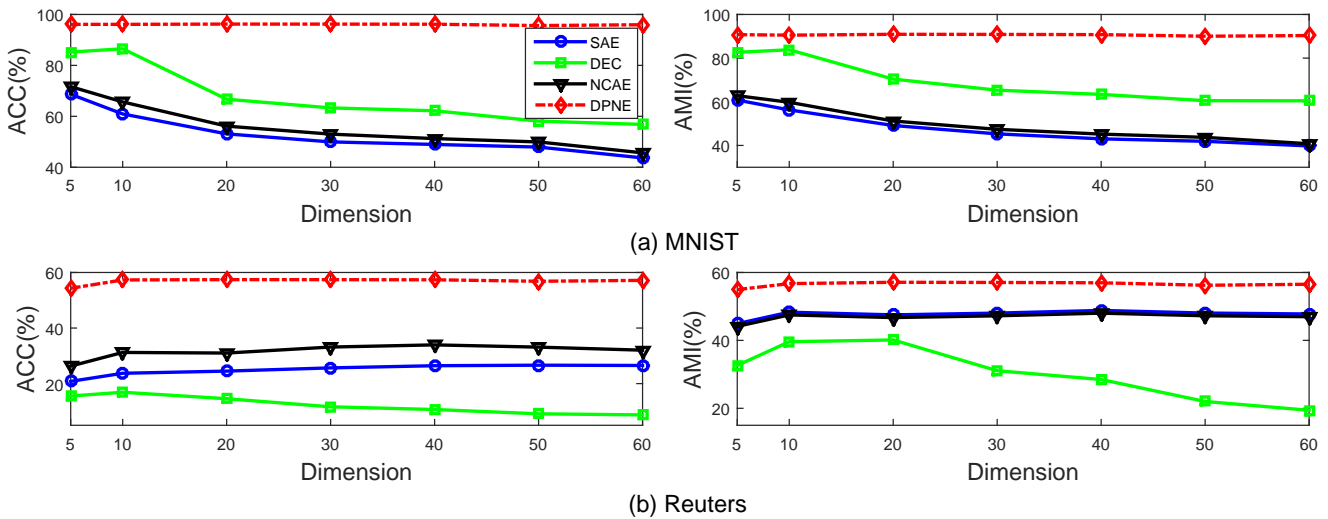}}
\caption{Dimension of the embedding space. (a) MNIST (b) Reuters.}
\label{Dimension}
\end{figure*}

\begin{figure*}[t]
\centerline{\includegraphics[width=16cm,height=6.2cm]{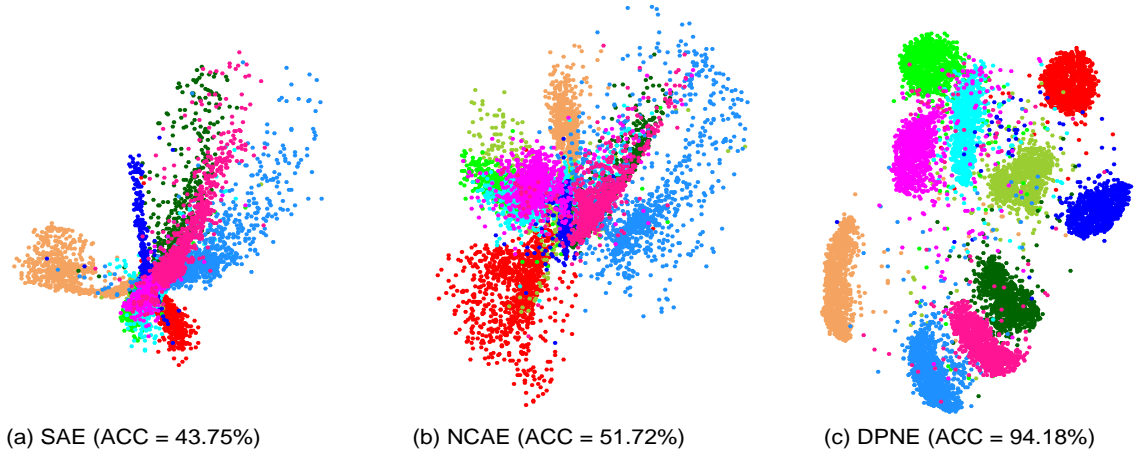}}
\caption{Visualize the 2-dimensional representation space obtained by (a) SAE, (b) NCAE and (c) DPNE.}
\label{Visualize2D}
\end{figure*}

\begin{figure*}[t]
\centerline{\includegraphics[width=18cm,height=9.5cm]{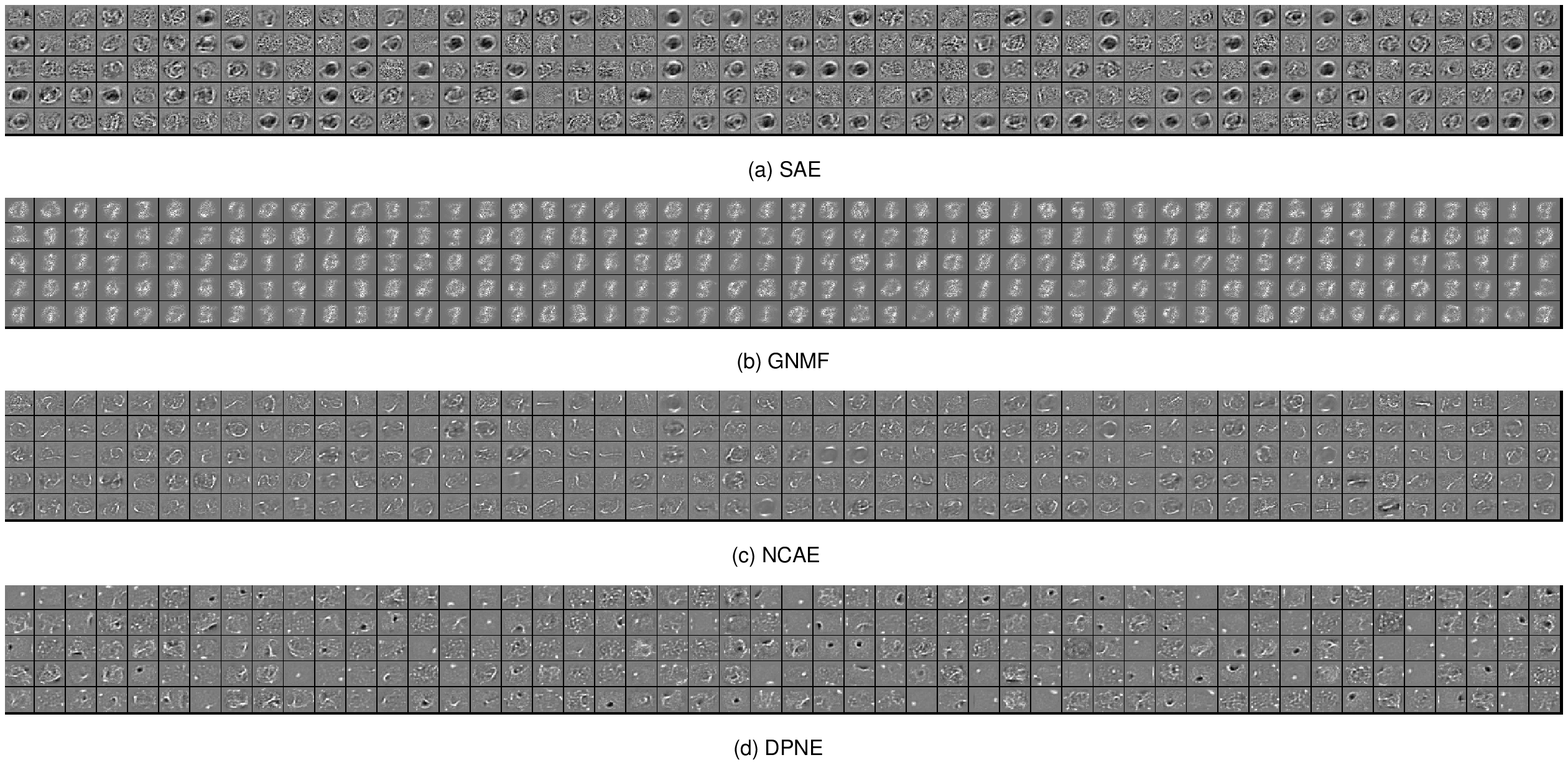}}
\caption{Randomly selected fine-tuning receptive fields of 250 out of 500 neurons from the first layer on MNIST data set. (a) SAE, (b) GNMF, (c) NCAE and (d) DPNE. Black pixels represent negative weights, gray pixels represent $\omega_{ij}^{(1)} = 0$ and white pixels represent positive weights.}
\label{W_display}
\end{figure*}

\begin{figure*}[t]
\centerline{\includegraphics[width=18cm,height=9.5cm]{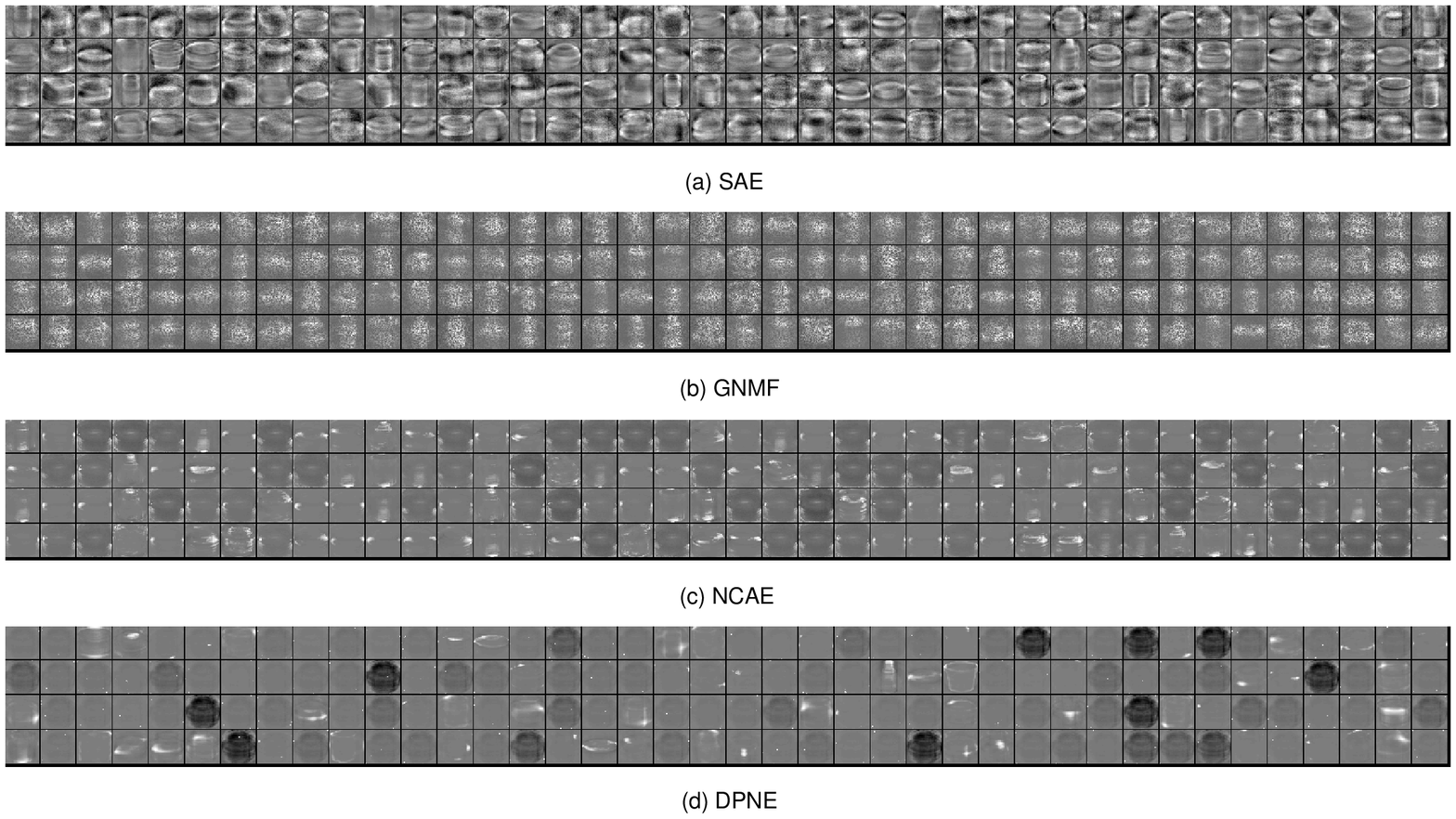}}
\caption{Randomly selected fine-tuning receptive fields of 160 out of 500 neurons from the first layer on the COIL100 data set. (a) SAE, (b) GNMF, (c) NCAE and (d) DPNE. Black pixels represent negative weights, gray pixels represent $\omega_{ij}^{(1)} = 0$ and white pixels represent positive weights.}
\label{W_display1}
\end{figure*}

\begin{table*}[t]
\caption{Clustering results (\%) of different methods.}
\centering
{\begin{tabular}{cccccc|ccccc} \toprule
    &\multicolumn{5}{c|}{ACC}   & \multicolumn{5}{|c}{AMI} \\ \hline
          & MNIST & Coil100 & YaleB & Reuters & RCV1 & MNIST & Coil100 & YaleB & Reuters & RCV1 \\
k-means++ &  53.2 & 62.1   &  51.4  &  23.6  &  52.9  & 50.0  &  80.3  &  61.5  &  51.6  & 35.5 \\
LDMGI    &  72.3  & 76.3   &  90.1  &  46.5  &  66.7  & 76.1   &  88.8  &  94.5  &  52.3   & 38.2  \\
GDL      &  n/a  & 82.5   &  78.3  &  46.3  &  44.4  & n/a   &\bfseries 95.8  &  92.4  &  40.1   & 2.0  \\
SEC       &  54.5 & 64.8   &  72.1  &  43.4  &  42.5  & 46.9  &  84.9  &  84.9  &  49.8  & 6.9  \\
DEC       &  86.7 & 81.5   &  2.7   &  16.8  &  68.3  & 84.0  &  61.1  &  0.0   &  39.7  &\bfseries 50.0 \\
DCN       &  56.0 & 62.0   &  43.0  &  22.0  &\bfseries 73.0  & 57.0  &  81.0  &  59.0  &  43.0  & 47.0 \\
RCC      &  87.6 & 83.1   &  93.9  &  38.1  &  35.6  & 89.3  &  95.7  &  97.5  &  55.6  & 13.8 \\
SAE       &  60.9 &  54.8  &  47.5  & 23.7   &  49.1  & 56.3 & 32.5 & 62.7 & 48.3  & 43.8 \\
NCAE      &  65.6  & 57.9   & 50.9   & 31.2   &  50.3  &  59.8 & 34.8 & 65.1 &  47.5  & 44.9 \\
DPNE      &\bfseries 96.1 &\bfseries 85.0  &\bfseries 94.8 &\bfseries 57.3  &  56.2 &\bfseries 90.5 &\bfseries 95.8 &\bfseries 98.5 &\bfseries 56.7 &\bfseries 50.0 \\ \toprule
\end{tabular}}
\label{ACC_AMI}
\end{table*}

We evaluated our proposed method against five widely used data sets: MNIST, Coil-100, YaleB, Reuters21578 and RCV1. Three of them are image data sets, and the rest are text data sets. The MNIST is the classic data set of 70000 hand-written digits, and each digit is a $28\times 28$ gray scale image \cite{Lecun1998Gra}. The Coil-100 data set contains $128\times 128$ color images of 100 objects viewed from different angles \cite{Nene1996Columbia}. The YaleB data set contains images of faces of 38 human subjects, and each one is a $192\times 168$ gray scale image \cite{Georghiades2001From}. The Reuters21578 and RCV1 are well-known data sets for text classification \cite{Lewis2004RCV1}. SVHN?

\subsection{Parameters Selection}
First, approximating the real distribution is crucial for the proposed DPNE. The number of nearest neighbors $k$ is a essential parameter for the density estimation of original high dimensional space. In this experiment, we restrict the proposed DPNE to a 10-dimensional representation space, i.e., $D = 10$. The average performance (ACC and AMI) of DPNE varies with the parameter $k$ is shown in Fig.~\ref{nearest_neighbor}.

It is observed from the Fig.~\ref{nearest_neighbor} that the clustering performance of DPNE is very stable with respect to the number of nearest neighbors $k$. When the parameter $k$ varies from 10 to 15, the DPNE can achieve better clustering performance. As we have described above, the DPNE employs the $k$-nearest neighbor kernel density estimation to approximate the manifold structure of original data. The parameter $k$ controls the width of high density area, where the data points belong to the same cluster. A too small $k$ will decrease the width of high density area and influence the the performance of density estimation. A too large $k$ will increase the width of high density area and be bad for capturing the manifold structure embedded in the high dimensional data space. So in this work, we set the number of nearest neighbors $k$ to 10.

Next, another experiment is to verify that the DPNE is robustness to dimensional $D$ of the latent feature space. The dimensional $D$ is varied between 5 and 60, and the clustering performance on the MNIST and the Reuters data sets are shown in Fig.~\ref{Dimension}. For comparison, the clustering performance of SAE, NCAE and DEC are also reported in the Fig.~\ref{Dimension}.

As we can see from the Fig.~\ref{Dimension}, the performance of SAE, NCAE and DEC on MNIST data set gradually decrease as parameter $D$ increases. The result supports the common view that clustering is more likely to fail as the dimensional $D$ increases. As what we have hoped, the performance of DPNE is very robust with respect to the dimensional $D$ of latent feature space. The DPNE achieves consistently good performance when dimensional $D$ changes from 5 to 60. For example, on MNIST data set, when the dimensional $D$ varies from 5 to 60, the ACC and AMI of the proposed DPNE only drop by $0.16\%$ and $0.46\%$, respectively. It is obvious that the part-based representation obtained by the DPNE model successfully retains the manifold structure embedded in the high dimensional data space. Then the learned feature becomes robust and is good for clustering and classification tasks.

\subsection{Visualization}
In this section, when the high dimensional data is embedded into a 2-dimensional plane, we analyze the outputs of SAE, NCAE and DPNE, respectively. We use the randomly sampled subset of 10000 observations from the MNIST data set to verify this purpose. We visualize the 2-dimensional feature space obtained by the SAE, NCAE and the proposed DPNE, as shown in Fig.~\ref{Visualize2D}.

As we can see, the proposed DPNE can reorganize the location of each sample in the 2-dimensional plane according to the original data distribution, and these locations successfully reveal the manifold structure of the original data. The 2-dimensional representation produced by the DPNE is well organized to cluster these samples. We also provide the corresponding cluster accuracy. Compare to the SAE and NCAE, it is observed that preserving the distribution of original data can significantly improve the cluster accuracy.

\subsection{Learning Part-based Representation}
Fig.~\ref{W_display} and Fig.~\ref{W_display1} show the weights vectors learned by the SAE, GNMF, NCAE and DPNE in the MNIST and COIL100 data sets respectively. Each weights vector of MNIST data set has dimensionality 784 and we plot the vector as $28\times 28$ gray image. The dimensional of weights vector of COIL100 data set is 16384 and the cropped gray image ($32\times 32$) is plotted.

As indicated in Fig.~\ref{W_display} (b) and Fig.~\ref{W_display1} (b), even through the receptive fields obtained by GNMF are all constrained to be non-negative, it is clear to see that the weights are much less sparse. The receptive fields learned from the MNIST data set by SAE does not have visually interpretable, and the weights learned from COIL100 data set indicate holistic features from different objects, as shown in Fig.~\ref{W_display} (a) and Fig.~\ref{W_display1} (a). Compared to SAE, the non-negative constrain deep models (i.e., NCAE and DPNE) have fewer darker pixels in the visualization of receptive fields. It can be seen from Fig.~\ref{W_display} (c), (d) and Fig.~\ref{W_display1} (c), (d) that the basic components of the samples can be discovered by the NCAE and DPNE methods. So the non-negative constrain deep models, NCAE and DPNE also can learn a part-based representation from the high dimensional observation.

\subsection{Comparisons with Other Algorithms}
Table \ref{ACC_AMI} shows the clustering results (accuracy and adjusted mutual information) of different methods. As in the case of DEC and DCN models, the dimension of the low dimensional representation (SAE, NCAE and the proposed DPNE) is set to be 10, i.e., $D = 10$. Compared to k-means, k-means++ can dramatically improve both the speed and accuracy \cite{Arthur2007k}. So we directly employ the k-means++ to cluster the original data without dimensionality reduction. Due to the size of the the full MNIST data set, the GDL method fails to yield a result and the clustering results are marked as 'n/a'. As indicated in Table \ref{ACC_AMI}, the performance of RCC is consistent on image data sets, but fails to cluster the text data sets. The deep models, DEC, DCN, SAE and NCAE that are based on autoencoder, also do not achieve consistent accuracy.

It is observed from Table \ref{ACC_AMI} that the AMI of our DPNE is consistent on the five data sets and the proposed DPNE can achieve the highest accuracy on four of the five data sets. This suggests the importance of inherent structure in learning the low dimensional representation. We also observe that the NCAE outperforms the SAE, which means the superiority of the meaningful representation idea in extracting the hidden structure. By leveraging the superiority of both the part-based representation and distribution preserving, the proposed DPNE can learn a better discriminative feature.

\section{Conclusion}\label{Conclusion}
In this paper, we present a novel dimensionality reduction method, called distribution preserving network embedding (DPNE). In DPNE, we use the data distribution to approximate the latent geometrical structure embedded in the high dimensional data space. And then the proposed DPNE utilizes the superiority of the part-based representation to learn a meaningful feature which respects to the above distribution. From analysis of the visualization results, the 2-dimensional part-based representation space obtained by the DPHE preserves the structure buried in the original high dimensional data as much as possible. The experimental results on the image and text data sets, also show that the proposed DPNE can learn a low dimensional part-based feature that has more discriminating power.

% if have a single appendix:
%\appendix[Proof of the Zonklar Equations]
% or
%\appendix  % for no appendix heading
% do not use \section anymore after \appendix, only \section*
% is possibly needed

% use appendices with more than one appendix
% then use \section to start each appendix
% you must declare a \section before using any
% \subsection or using \label (\appendices by itself
% starts a section numbered zero.)
%

%\appendices
%\section{Proof of the First Zonklar Equation}
%Appendix one text goes here.

% you can choose not to have a title for an appendix
% if you want by leaving the argument blank
%\section{}
%Appendix two text goes here.

% use section* for acknowledgment
\ifCLASSOPTIONcompsoc
  % The Computer Society usually uses the plural form
  \section*{Acknowledgments}
\else
  % regular IEEE prefers the singular form
  \section*{Acknowledgment}
\fi

The authors would like to thank the handling editor and anonymous reviewers for their insights and comments in helping improve our work.

% Can use something like this to put references on a page
% by themselves when using endfloat and the captionsoff option.
\ifCLASSOPTIONcaptionsoff
  \newpage
\fi

% trigger a \newpage just before the given reference
% number - used to balance the columns on the last page
% adjust value as needed - may need to be readjusted if
% the document is modified later
%\IEEEtriggeratref{8}
% The "triggered" command can be changed if desired:
%\IEEEtriggercmd{\enlargethispage{-5in}}

% references section

% can use a bibliography generated by BibTeX as a .bbl file
% BibTeX documentation can be easily obtained at:
% http://mirror.ctan.org/biblio/bibtex/contrib/doc/
% The IEEEtran BibTeX style support page is at:
% http://www.michaelshell.org/tex/ieeetran/bibtex/
%\bibliographystyle{IEEEtran}
% argument is your BibTeX string definitions and bibliography database(s)
%\bibliography{IEEEabrv,../bib/paper}
%
% <OR> manually copy in the resultant .bbl file
% set second argument of \begin to the number of references
% (used to reserve space for the reference number labels box)
%\begin{thebibliography}{1}
%
%\bibitem{IEEEhowto:kopka}
%H.~Kopka and P.~W. Daly, \emph{A Guide to \LaTeX}, 3rd~ed.\hskip 1em plus
%  0.5em minus 0.4em\relax Harlow, England: Addison-Wesley, 1999.
%
%\end{thebibliography}

\bibliographystyle{IEEEtran}
\bibliography{IEEEabrv,DPNE}

% biography section
%
% If you have an EPS/PDF photo (graphicx package needed) extra braces are
% needed around the contents of the optional argument to biography to prevent
% the LaTeX parser from getting confused when it sees the complicated
% \includegraphics command within an optional argument. (You could create
% your own custom macro containing the \includegraphics command to make things
% simpler here.)
%\begin{IEEEbiography}[{\includegraphics[width=1in,height=1.25in,clip,keepaspectratio]{mshell}}]{Michael Shell}
% or if you just want to reserve a space for a photo:

% You can push biographies down or up by placing
% a \vfill before or after them. The appropriate
% use of \vfill depends on what kind of text is
% on the last page and whether or not the columns
% are being equalized.

%\vfill

% Can be used to pull up biographies so that the bottom of the last one
% is flush with the other column.
%\enlargethispage{-5in}

% that's all folks
\end{document}